\documentclass[conference]{IEEEtran}
\IEEEoverridecommandlockouts
\usepackage[T1]{fontenc}
\usepackage{url}
\usepackage{cite}
\usepackage{amsmath,amssymb,amsfonts}
\usepackage{algorithmic}
\usepackage{graphicx}
\usepackage{textcomp}
\usepackage{xcolor}
\usepackage[most]{tcolorbox}
\usepackage{mdframed}
\usepackage{lipsum}
\usepackage{wrapfig}

\mdfsetup{font=\small}
\mdfsetup{linewidth=1pt}

\def\BibTeX{{\rm B\kern-.05em{\sc i\kern-.025em b}\kern-.08em
    T\kern-.1667em\lower.7ex\hbox{E}\kern-.125emX}}
\begin{document}
\pagestyle{empty}
\pagestyle{plain}
\title{Bridging Natural Language and ASP: A Hybrid Approach Using LLMs and AMR Parsing \\
\thanks{Approved for public release: distribution is unlimited. Case Number AFRL-2025-1920. The views expressed are those of the authors and do not reflect the official guidance or position of the United States Government, the Department of Defense, or the United States Air Force.}
}

\author{Connar Hite, Sean Saud, Raef Taha, Nayim Rahman, Tanvir Atahary, Scott Douglass, and Tarek Taha \\
hitec2@udayton.edu, sauds2@udayton.edu, tahar1@udayton.edu, rahmanm12@udayton.edu, \\ataharyt1@udayton.edu, scott.douglass.1@us.af.mil, and tarek.taha@udayton.edu
}

\maketitle

\begin{abstract}
Answer Set Programming (ASP) is a declarative programming paradigm based on logic programming and non-monotonic reasoning. It is a tremendously powerful tool for describing and solving combinatorial problems. Like any other language, ASP requires users to learn how it works and the syntax involved. It is becoming increasingly required for those unfamiliar with programming languages to interact with code. This paper proposes a novel method of translating unconstrained English into ASP programs for logic puzzles using an LLM and Abstract Meaning Representation (AMR) graphs. Everything from ASP rules, facts, and constraints is generated to fully represent and solve the desired problem. Example logic puzzles are used to demonstrate the capabilities of the system. While most current methods rely entirely on an LLM, our system minimizes the role of the LLM only to complete straightforward tasks. The LLM is used to simplify natural language sentences, identify keywords, and generate simple facts. The AMR graphs are then parsed from simplified language and used to generate ASP constraints systematically. The system successfully creates an entire ASP program that solves a combinatorial logic problem. This approach is a significant first step in creating a lighter-weight, explainable system that converts natural language to solve complex logic problems.
\end{abstract}


\section{Introduction}
Translation of language from one form to another is often a lengthy but necessary endeavor. This fact has not changed with the rise and expanded use of computers. It is becoming increasingly required for those unfamiliar with programming languages to interact with code. Thus, there has been an increased focus on developing systems to convert natural language into logic programs. One area that has been neglected in the past is Knowledge Representation and Reasoning (KRR). For the sake of our work, the KRR formalism being generated is Answer Set Programming (ASP) \cite{2012GebserAnswerSetSolving}. While groups are developing means for the automatic generation of KRR formalisms, much is still needed to create a flexible, robust, and easy-to-use tool for those unversed in that field.

Deep learning and Natural Language Understanding (NLU) have seen encouraging results when addressing the problem of generating code from natural language. Deep learning approaches often focus on leveraging Large Language Models (LLMs) and knowledge modules to support the LLM. This method proves to be faster to develop while still providing favorable results. However, hallucination and bias are often struggles encountered while developing LLM-based systems \cite{10.1145/3571730}. They also struggle with common-sense reasoning \cite{bommasani2022opportunitiesrisksfoundationmodels}, with knowledge modules often used to mitigate this problem. Knowledge modules by no means completely solve the issue and are often handcrafted for specific problems, making them very time-consuming to create. The results of these methods suffer from a lack of explainability. Progress has been made in this area; however, LLMs still prove to be a black box when it comes to explaining the reasoning behind the generated answer \cite{bommasani2022opportunitiesrisksfoundationmodels}. Ishay et al. \cite{ishay2023leveraginglargelanguagemodels} utilize LLMs to generate ASP code for logical puzzle-solving. They separate the generation of ASP code into multiple prompts. Using only few-shot learning examples in their prompts, they can generate reasonably complex ASP programs. Similarly, Pan et al. \cite{pan-etal-2023-logic} introduced Logic-LM, which combines LLMs and symbolic solvers to solve logic problems. While effective, both methods suffer from several of the common pitfalls of LLMs previously discussed. An LLM's knowledge is limited to the training data that may not cover all ASP methods and syntax, so it may not produce accurate ASP rules for complex logic.

The field of NLU seeks to address this issue by automatically extracting the underlying meaning behind text and utilizing that to generate code. These methods take much effort to develop and tend to be more inflexible. However, the result is a structure that is consistent and capable of generating more complex outputs. In \cite{CARUSO_DODARO_MARATEA_MOCHI_RICCIO_2024}, a system using Controlled Natural Language (CNL) to generate ASP code is proposed. This system is expanded upon in \cite{borroto2024automaticcompositionaspprograms} to allow for natural language inputs. This is accomplished by converting natural language into controlled natural language.

This work proposes a novel model for generating ASP code for combinatorial logic puzzles from natural language. LLMs are used to capture the information regarding the problem space and to structure the constraints found in the problem. Their role has been minimized to avoid some of the pitfalls associated with their use. Compared to solely LLM-based systems, our proposed system guarantees correct syntax given the proper inputs. AMR graphs are combined with this information to generate the necessary constraints for the problem. These are used due to their ability to represent the core meaning behind a sentence. While the AMR parser in this system is LLM-based, it does not directly generate ASP code. The result is a runnable ASP program that represents the totality of the original problem. This approach is a significant first step in creating a lighter-weight, more explainable system that converts natural language to solve complex logic problems.

\section{Background}
\subsection{Answer Set Programming}\label{AA}
Answer Set Programming (ASP) is a declarative programming paradigm based on logic programming and non-monotonic reasoning. It is widely used in knowledge representation and combinatorial problem-solving \cite{2012GebserAnswerSetSolving}. It can express and solve NP-complete problems, making it very popular in the deep learning field \cite{10.1145/2043174.2043195}. It is similar to First-Order Logic (FOL) in how it expresses formal logic problems. However, ASP is more focused on efficiently solving a subset of said problems. ASP code describes what a solution should look like rather than describing how to find a solution \cite{2012GebserAnswerSetSolving}. From a high-level perspective, ASP considers all possible solutions and then narrows them down based on further considerations. These problems can be broken into two general steps: problem instance and problem encoding. Both steps are made of rules that establish and solve a problem. For the purpose of this paper, Clingo is the system used as the grounder and solver. Therefore, the generated ASP is in a format that Clingo accepts. Normal rules are in the form:
\begin{equation}
P_0 :-\ P_1, ..., P_m.
\end{equation}

Where P\textsubscript{0} is known as the head and P\textsubscript{1}, ... ,P\textsubscript{m} is the body of the rule. This can be understood as the head is true if the body holds true. If there is no body, then the head is taken as always true. This is known as a fact in ASP. The problem instance comprises facts that establish what is known about the problem space \cite{2012GebserAnswerSetSolving}. For example, this could be the names of people:

\begin{mdframed}\textit{\noindent
names("John"; "Nick"; "Sydney"; "Teri").
}
\end{mdframed}

The predicate \textit{names} is used to indicate what group a constant belongs to. In ASP, constants are strings typically beginning with a lowercase letter, while variables begin with a capital letter. If a constant must start with a capital, such as "John", then it must be enclosed in quotes to distinguish it from a variable. Variables serve as placeholders within a rule and are grounded during the solving process. Other important components found within rules are the logical relationships represented by commas, signifying an "and" relationship, and semicolons, signifying an "or" relationship. These components are all used within the problem encoding section of an ASP problem to define relationships, set constraints, and generate potential solutions. The generation rules are a key part of this, as they describe the form of all possible solutions \cite{2012GebserAnswerSetSolving}. Beyond the basic rule format, another common structure is the choice rule, which has the following format:
\begin{equation}
X\{P_1, P_2, ..., P_m : P_n\}Y :-\ P_{m+1}.
\end{equation}

The brackets indicate what choices can be made. The left side of the \textit{if} statement is known as the cardinality constraint. X and Y indicate the lower and upper bounds, respectively. If left blank, then no bounds are set. The cardinality constraint is satisfied when the number of true literals, atoms, is between the lower and upper bounds. This statement could be read as: Given P\textsubscript{m+1}, between X and Y of the atoms must be true, P\textsubscript{m}. P\textsubscript{n} is used to provide more information about the choice. Choice rules allow for specifying that some atoms may or may not be true without requiring a deterministic decision \cite{2012GebserAnswerSetSolving}. For example, the first line in the following equation:
\begin{equation}
\begin{gathered}
\{a, b\} :-\ C.\\
\{a, b\}, \{a\}, \{b\}, \{\}.
\end{gathered}
\end{equation}

results in the stable models found in the second line (assuming c is true). This means that any combination of atoms a and b may be valid in the answer set. Constraints have the same format as rules but have no head. These are used while solving to eliminate possible solutions to the problem \cite{2012GebserAnswerSetSolving}. An example could be getting rid of specific solutions:

\begin{mdframed}\textit{\noindent
:- solution("John", wednesday, 1).
}
\end{mdframed}

This line would remove only that specific solution. Variables and multiple literals can be used to remove broader groups. Lastly, some information is passed to the grounder to identify what is being solved. For our case, these lines appear at the end of the ASP file and follow this format:

\begin{mdframed}\textit{\noindent
\#hide.\\
\#show P\textsubscript{1}/n.
}
\end{mdframed}

The first line hides all solutions in the output, while the second includes all solutions for predicate P1/n. Including these lines creates a more readable solution.

\subsection{Large Language Models}
Large Language Models (LLMs) are often employed when generating ASP code and performing Natural Language Processing (NLP) tasks \cite{10.1145/3605943}. LLMs are complex machine learning systems that can understand, generate, and manipulate human language. They are built primarily using transformer-based architectures and trained on vast amounts of data. They can be fine-tuned for specific tasks, making them versatile in complex tasks such as natural language processing, content creation, code generation, summarizing text, answering questions, and much more \cite{10.1145/3605943}. Prompts to LLMs need to be engineered to generate the most efficacious responses. There are multiple methods to prompt engineering, but most aim to mimic human reasoning. Ideal prompts generally need to guide the LLM while being clear and concise.

There are an assortment of LLMs to choose from for any given problem. In this paper, Chat-GPT 4o is used. It is one of the larger, publicly available LLMs. Different versions of Llama have been tested for a smaller model; however, the results need to be developed further.

\subsection{Abstract Meaning Representation}
\begin{figure}[htbp]
\centerline{\includegraphics[width=.23\textwidth]{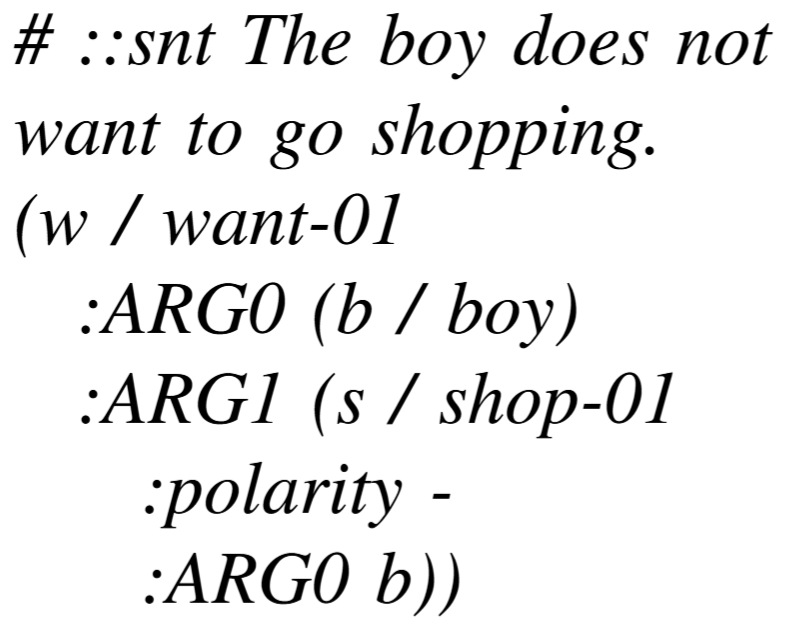}
\includegraphics[width=.23\textwidth]{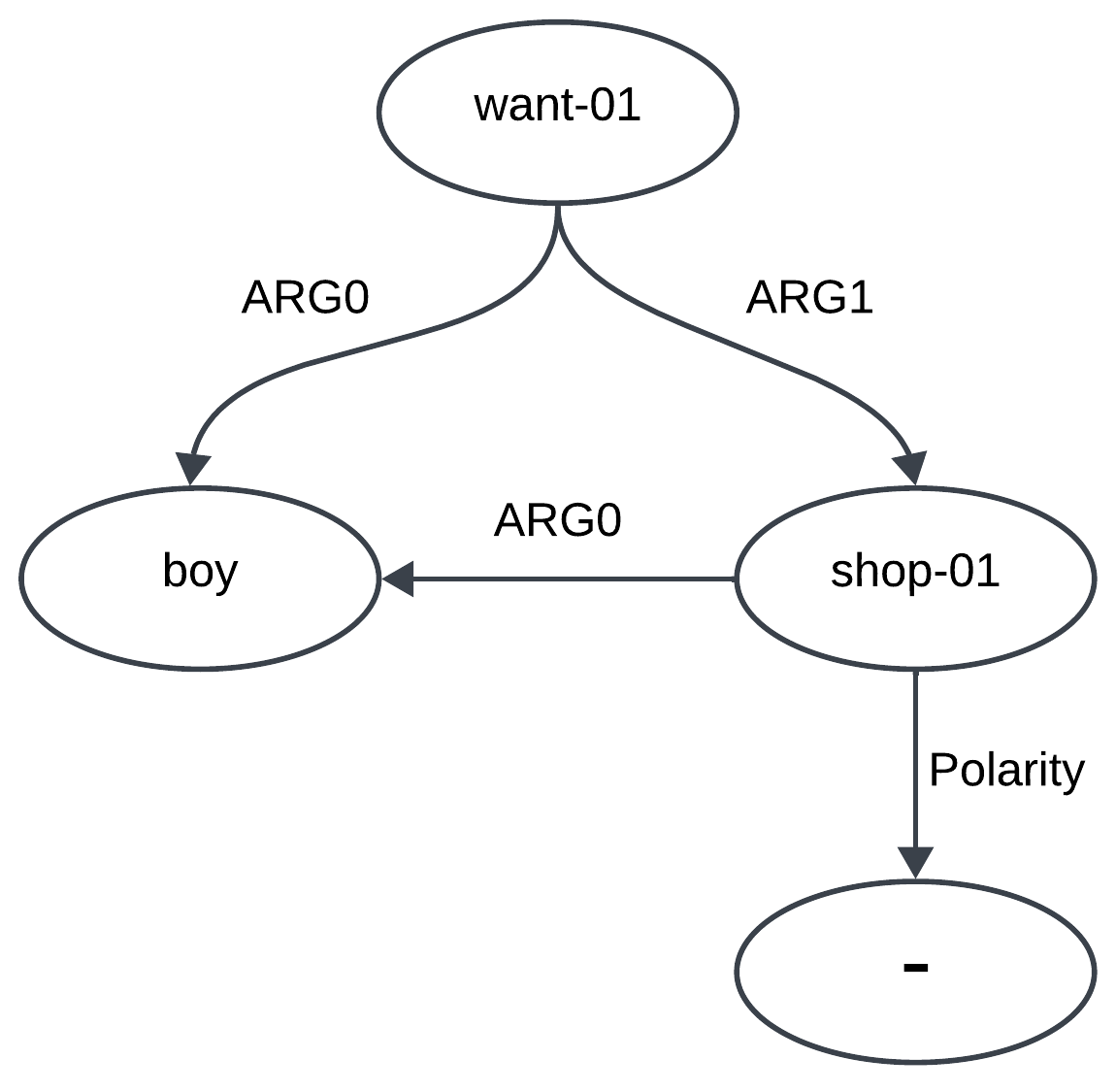}}
\caption{Example AMR Graph}
\label{fig:ex_amr}
\end{figure}

Abstract Meaning Representation (AMR) is a semantic representation language used in natural language processing \cite{banarescu-etal-2013-abstract}. More specifically, AMR is used in natural language understanding, natural language generation, and statistical machine translation \cite{tosik2024abstractmeaningrepresentation}. The general purpose is to capture the meaning of sentences, regardless of their form, which is the core idea behind the abstract nature of AMR. A result of this approach is that each word in the original sentence does not result in a node in the AMR. Instead, only the key ideas are translated into nodes. The two standard methods used to represent an AMR are PENMAN notation, which is used in this paper, and conjunctions of logical triples. PENMAN notation is a serialized format for directed, rooted graphs. Each node in the graph represents a concept from the original sentence. These can be words, PropBank frames, or special keywords. Roles and variables are used to describe how concepts connect and their contribution to the original sentence. An example sentence is shown in Figure \ref{fig:ex_amr}.

Nodes such as "shop-01" are selected from the PropBank library. This repository contains frames used to structure the semantic representation for the meaning of words and their arguments. If applicable, a single word may also have multiple meanings, indicated by the number to the right. For example, the word "flop" has two definitions: to fall/fail or ungainly movement. The former would be marked as "flop01", ARG0 would indicate what caused the event, and ARG1 would be what flopped. This results in an AMR graph representing a sentence's core meaning. For the above example, ARG0 (boy) is the wanter and ARG1 (shop) is the thing wanted. However, the polarity under ARG1 indicates that it is not wanted. The ARG0 (b) of shop indicates the shopper. The \textit{b} is used to indicate that a previously defined node (boy) is being referenced. Altogether, the AMR breaks down who/what is being talked about, their status, and how each concept connects to one another. One of the benefits is that even if the language used is different, the resulting graph should be the same.

Once an AMR is generated, it can be used to parse concepts, generate new sentences, and analyze how it compares to another AMR graph \cite{tosik2024abstractmeaningrepresentation}. For the case of this paper, we are interested in parsing out certain concepts found within the AMR. This is done by looking at what nodes are contained within the AMR and how they are connected. Sentence generation is a field of its own. It is accomplished by piecing the sentence back together based on concepts and the meaning of each argument. The resulting sentence, in most cases, differs from the original sentence. However, the core meaning should still be there. Lastly, AMR comparison tests how similar two AMRs are based on their layout. This is most commonly done using a score based on matching edges in each graph. During training, this method is used to test the accuracy of a system. The comparison is often made between hand-made, ideal AMRs and machine-generated ones.

\section{Related Works}
Recently, there has been a significant focus on code generation, with most approaches focusing on a specific language. The prospect of automatically generating ASP code is no different, and several attempts have been made. The most common approach is the use of LLMs. The LLM can be used to parse and format the input into usable code. This technique has been used to generate first-order logic to solve a precise task, such as claim verification \cite{wang-shu-2023-explainable} or statement validity \cite{Olausson_2023}. The inputs are converted into FOL, which can then be processed accordingly. Rajasekharan et al. \cite{Rajasekharan_2023} use a similar method in their STAR system to generate ASP code. However, their focus is on solving NLU tasks with the help of common sense modules. Each of these methods is highly geared towards its specific task. 

In contrast, Coppolillo et al. \cite{coppolillo2024llaspfinetuninglargelanguage} focus on constructing ASP code through a specialized LLM. Each function has a corresponding template used to create the desired concept in ASP, allowing for some flexibility, but it is limited to existing templates. The framework in \cite{ishay2023leveraginglargelanguagemodels} utilizes LLMs to generate code for logical puzzle-solving. The generation of ASP code is separated into multiple prompts. First, constants are extracted, then predicates are generated, and finally, rules are generated. They can generate fairly complex ASP programs using few-shot learning examples in their prompts. Few-shot learning with LLMs is also utilized in \cite{li2025logic} to bridge the gap with logic programming, but frames it as a broader "Logic-of-thought" paradigm and has near-perfect performance. 

Logic-LM \cite{pan-etal-2023-logic} combines LLMs and symbolic solvers to solve logic problems and shows significantly better performance than using an LLM alone. It includes a self-refinement feature in which the solver's error messages are used to revise symbolic formulations that the LLM writes. The system proposed in \cite{wang2024dspybasedneuralsymbolicpipelineenhance} is inspired by Logic-LM but focuses on enhancing spatial reasoning. The proposed system incorporates DSPy for iterative feedback between the LLM and ASP. Kaur et al. \cite{kaur2025empirical} introduce conformal prediction to the LLM-to-ASP process. In the proposed system, an LLM generates a set of ASP programs and then selects the strongest ASP program based on a novel metric. This increases robustness, but creates substantial overhead. 

Additionally, Learning from Answer Sets (LAS) can be implemented with LLMs to learn ASP rules by comparing the outputs of ASP solvers with target answer sets \cite{borroto2025questionansweringllmslearning}. This improves consistency by allowing structured correction of LLM-generated ASP code, but requires supervision from labeled answer sets. These LLM-based systems generally suffer from issues with integer constant formatting, assuming incorrect information, syntax errors, and other logical errors. 

Another approach to generating ASP code involves Controlled Natural Language (CNL). The purpose of CNL is to restrict the structure and vocabulary of sentences to fit the objective better. Through parsing, ASP can then be generated from CNL \cite{10.1007/978-3-642-32612-7_3} \cite{schwitter2014controllednaturallanguageprocessing} \cite{SCHWITTER_2013}. CNL can also be used reversely to translate ASP into readable English \cite{schwitter2018specifyingverbalisinganswerset}. This idea is expanded on in \cite{CARUSO_DODARO_MARATEA_MOCHI_RICCIO_2024} by generating ASP from human-readable inputs, focusing on combinatorial problems. For each ASP concept implemented, there is a specific format that the sentence follows. This results in a predictable output that is solely based on the user. This idea is expanded further in \cite{borroto2024automaticcompositionaspprograms} by allowing natural language inputs. Their system, NL2ASP, utilizes Neural Machine Translation (NMT) to convert natural language into controlled natural language. The output is placed into the CNL2ASP system, which generates the ASP. CNLWizard \cite{carusogeneral} simplifies the process of creating a CNL. It allows automatic translation into multiple knowledge representation formalisms, including ASP. This significantly reduces the time and effort to create custom CNLs.

\section{Methodology}
\begin{figure}[htbp]
\centerline{\includegraphics[width=.5\textwidth]{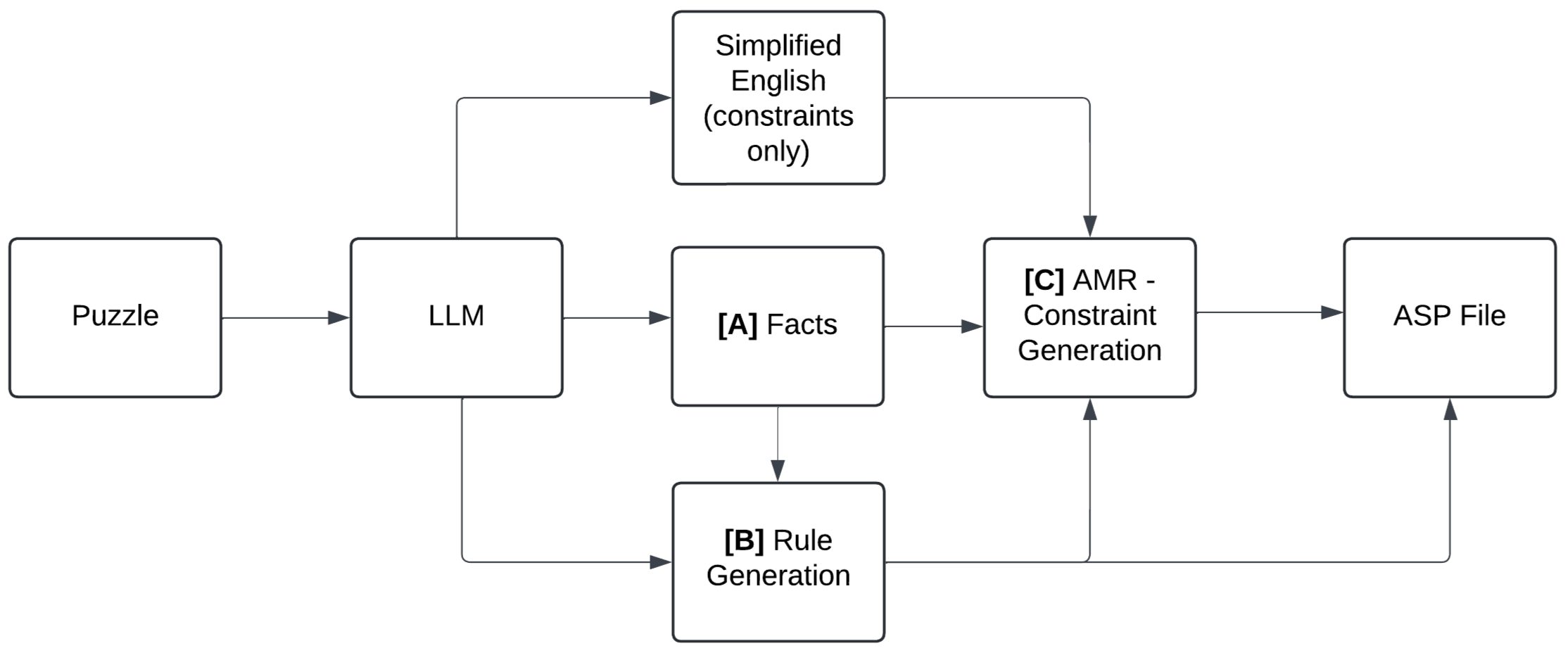}}
\caption{System Architecture}
\label{fig:sys_arch}
\end{figure}

Generating the entire ASP file in one step has proven impractical, necessitating a process that is broken down into multiple smaller steps. This decomposition aligns with the human reasoning process typically used when solving ASP problems by hand. Figure \ref{fig:sys_arch} outlines the structure of our system and the processes found within it. The first step is determining what is known, or the facts, of the problem space \textbf{[A]}. Once this information has been established, the relationships, or the rules, can be created. During this process, the format of the final solution can also be set \textbf{[B]}. Separately, the constraints found in the original problem can be made uniform and simplified \textbf{[C]}. Each of these steps utilizes the same LLM structure with different prompts. Doing so allows for information to be found throughout the problem without requiring the text to be heavily formatted. The final step is generating the ASP constraints using information generated in the previous steps and analyzing AMR graphs made from each constraint. Example problems are used to explain our process, which can be found in the appendix.

\subsection{Fact Generation}
Categories are extracted from the given problem statement using the following prompt for fact generation. The \textit{\{problem description\}} term is a placeholder that is replaced with the natural language description of the logic puzzle.

\begin{mdframed}\textit{\noindent
List each specific category of entities in this logic puzzle is asking to be determined, both implicitly and explicitly. Respond with only the list of categories and nothing else. Format the list in one line separated by commas. If multiple words are required to describe the category, separate them by underscores. Use only lowercase and alphanumeric characters. If a category in your list has a singular form, use the singular form. Take your time and double-check your answer. Use a value of 0 for the temperature.\\[0.3ex]
Problem description:
\{problem description\}
}
\end{mdframed}

Prompting LLMs to take their time has proven to increase the efficacy of their responses \cite{snell2024scalingllmtesttimecompute}. As ChatGPT does not allow for the adjustment of hyperparameters through its online interface, it is explicitly told to use a value of 0 for temperature to minimize randomness in responses. It is prompted to respond only with categories and nothing else, so it knows precisely what to output. This approach helps avoid a response containing unnecessary information (such as describing what it is responding to). An example problem used throughout this section is found in Appendix Figure 1. The result of running this prompt on said problem is as follows:

\begin{mdframed}\textit{\noindent
child,
favorite\_animal,
balloon\_design,
order\_in\_line,
gender
}
\end{mdframed}

These serve as the predicates for the facts. Additional prompting is required to craft the rest of the facts. The following prompt extracts the entities in the logic puzzle:

\begin{mdframed}\textit{\noindent
List all the entities in this logic puzzle. Respond with only the list of entities and nothing else. Format the list in one line separated by commas. If multiple words are required to describe the entity, separate them by underscores. Use only lowercase and alphanumeric characters. Take your time and double-check your answer. Use a value of 0 for the temperature.\\[0.3ex]
Problem description:
\{problem description\}
}
\end{mdframed}

The next prompt takes the categories and entities from the previous prompts as inputs and assigns each entity to a category. Including this as a separate step instead of directly generating the ASP facts from the categories and entities increased consistency in results.

\begin{mdframed}\textit{\noindent
Given the following categories and entities: \\
- Categories: \{categories from prompt1\}\\
- Entities: \{entities from prompt2\}\\[0.3ex]
Assign entities to each category to which they belong. Entities may belong to more than one category. Format the list of entities that belong to a certain category as follows:\\ 
category: entity\_1, entity\_2, …, entity\_N\\[0.3ex]
In the response, start a new line for each list of entities in a category. The response should only include the lists, and nothing else. Take your time and double-check your answer. Use a value of 0 for the temperature.\\[0.3ex]
Problem description: \{problem description\}
}
\end{mdframed}

Next, the associated categories and their entities are input to the following prompt to format them as ASP facts:

\begin{mdframed}\textit{\noindent
Given the following lists of entities assigned to a certain category:\\
\{response from prompt 3\}\\[0.3ex]
In terms of Answer Set Programming code, represent each list of entities assigned to a category as an ASP fact in the following format:\\
category("entitiy\_1";"entitiy\_2";...;"entitiy\_N").\\[0.3ex]
Start a new line after each ASP fact. Respond with only ASP facts and nothing else. If an entity is numeral or ordinal, represent it in numeral format without quotation marks. If a predicate (category), P1 is inherently an attribute of another predicate, P2, format a fact in the following format:\\
P1\_of("constant\_1, "constant\_2").\\[0.3ex]
Do not try to solve the logic puzzle. Take your time and double-check your answer. Use a temperature of 0.\\[0.3ex]
Problem description: \{problem description\}
}
\end{mdframed}

In order to extract additional facts, such as the gender of a child, the LLM is prompted to recognize when a predicate is inherently an attribute of another predicate and generate facts accordingly. The LLM must be explicitly prompted not to solve the entire problem to avoid the addition of extraneous information. This results in the generation of fully functional facts. For the investigated problem, the results are as follows:

\begin{mdframed}\textit{\noindent
child("Kerry";"Johan";"Mario";"Lani";"Naomi").\\
favorite\_animal("chimpanzees";"tigers";"zebras";\\"lions";"giraffes").\\
balloon\_design("rainbow";"hearts";"stripes";\\"swirls";"polka\_dots").\\
order\_in\_line(1;2;3;4;5).\\
gender("boy";"girl").\\
gender\_of("Johan","boy").\\
gender\_of("Mario","boy").\\
gender\_of("Kerry","girl").\\
gender\_of("Lani","girl").\\
gender\_of("Naomi","girl").
}
\end{mdframed}

\subsection{Rule Generation}
Rule generation is currently minimal due to the scope of problems having one-to-one pairings. Thus, the prompt for generating the required rules exclusively looks for such pairings. The following prompt is passed predicates extracted from the previous fact generation step:

\begin{mdframed}\textit{\noindent
Given predicates: \{predicates extracted\}\\
In the following prompt, which of the given predicates has a one-to-one pairing? Format the list in one line separated by commas. Use only lowercase and alphanumeric characters.
Take your time and double-check your answer. Use a temperature of 0:\\ 
\{problem description\}
}
\end{mdframed}

The predicates identified by the LLM are then passed to a program that uses them to create the necessary generative rules. Currently, a reference predicate is manually chosen to make more logically named rule names, but any predicate can be successfully used as a reference predicate. The program then loops through the predicates and formats them as choice rules, Equation 2. Applying this process to our chosen problem results in the following rules:

\begin{mdframed}\textit{\noindent
1\{favorite\_animal\_of(C,F):favorite\_animal(F)\}1:-child(C).\\
1\{favorite\_animal\_of(C,F):child(C)\}1:-favorite\_animal(F).\\
1\{balloon\_design\_of(C,B):balloon\_design(B)\}1:-child(C).\\
1\{balloon\_design\_of(C,B):child(C)\}1:-balloon\_design(B).\\
1\{order\_in\_line\_of(C,O):order\_in\_line(O)\}1:-child(C).\\
1\{order\_in\_line\_of(C,O):child(C)\}1:-order\_in\_line(O).
}
\end{mdframed}

At the same time, a single predicate is created to combine these into a final solution. Instructions for the grounder are also created using this predicate and the number of variables it contains. The result is as follows:

\begin{mdframed}\textit{\noindent
solution(Child,Favorite\_animal,Balloon\_design,\\Order\_in\_line):-favorite\_animal\_of(Child,\\Favorite\_animal),balloon\_design\_of(Child,\\Balloon\_design),order\_in\_line\_of(Child,\\Order\_in\_line).\\
\#show solution/4.
}
\end{mdframed}

\subsection{Constraint Generation}
Constraints are created last, as they depend on information generated in the preceding steps before compiling the final ASP file. Using a LLM, the constraints are separated and simplified. This allows each line to have a singular constraint. Additionally, the LLM is prompted with the previously extracted predicates and instructed to use them throughout. This ensures that each constraint uses consistent language. This is necessary as our system is currently not handling synonyms. If a line uses differing language, this could cause information to be lost. The same LLM in the previous steps is used with the following prompt:

\begin{mdframed}\textit{\noindent
Create a series of constraints from the following text. Never list constant/category relations as constraints. Never include irrelevant information (instructions, background, etc.). Output on separate lines; never use list format. Use these keywords throughout for consistent language: \{extracted predicates\}.\\
\{problem description\}
}
\end{mdframed}

The following text shows how the first constraint is simplified:

\begin{mdframed}\textit{\noindent
The girl whose favorite animal is the tigers was third in line.\\
Naomi's favorite animal is not the tigers.\\
The girl whose favorite animal is the tigers did not receive the balloon with hearts.\\
The girl whose favorite animal is the tigers did not receive the balloon with stripes.
}
\end{mdframed}

Our approach is reminiscent of chain-of-thought prompting from \cite{wei2023chainofthoughtpromptingelicitsreasoning}, which requires an LLM to use reasoning to explain how it arrives at an answer. This technique provides transparency regarding the model's reasoning while also improving model accuracy. Although the constraints are not being solved outright with an LLM, our approach mirrors chain-of-thought prompting by expanding the constraints and connecting details where relevant. The next step is to establish what is known about the problem space. What rules exist? What predicate is used, and what is the order of the variables? What do these variables correlate to? Lastly, what facts are established, and what constants are found there? All of these questions are found within the fact and rule generation sections. So, what is required is the extraction and formation of this information into a helpful structure. This assists in deciding what information in the AMR graph is pertinent to generating a constraint. If a constant is found within the AMR, then this is noted as a constant of interest. If part of a rule's predicate is found, then this is noted as a rule of interest. For example, this simplified constraint is used: "The girl whose favorite animal are tigers did not receive the balloon with hearts." This results in the following AMR graph:

\begin{mdframed}\textit{\noindent
\# ::snt The girl whose favorite animal is the tigers did not receive the balloon with hearts.\\
(r / receive-01\\
    \hspace*{3mm}:polarity -\\
    \hspace*{3mm}:ARG0 (g / girl)\\
        \hspace*{6mm}:ARG0-of (f / favor-01\\
            \hspace*{9mm}:ARG1 (t / tiger)))\\
    \hspace*{3mm}:ARG1 (b / balloon \\
        \hspace*{6mm}:ARG0-of (h / have-03\\
             \hspace*{9mm}:ARG1 (h2 / heart))))
}
\end{mdframed}

SpaCy \cite{honnibal2020spacy}, with the AMR extension found in \cite{Bjascob}, is used to generate the AMR graphs. The AMR model that is used is \textit{parse\_xfm\_bart\_large}. The nodes of this graph are compared to the known facts in order to identify important pieces of information. For this example, nodes "heart," "tiger," and "girl" would be noted. The same can be done for the rules, where the noted nodes are "balloon" and "favor." One important distinction between these areas of interest is how they are used. Facts are definite pieces of information that should be included. However, the identified rules might not be enough. The identified rules might not fully cover certain connections. In our example, the child's gender is not covered by the identified rules. How this is handled is explained later, but it is important to note.

With this, all the information required to construct the constraints is fully obtained. The two formats that are currently implemented are choice rules and integrity constraints. How the format is chosen depends on the complexity of the constraint. The variables and their polarities must be determined for each rule of interest found within a single constraint. The former is found by determining which fact corresponds to each variable within the rule. If a constant from said rule exists, it can be filled in. Otherwise, the variable is used and noted. Each variable should be used at least twice within a single line to corroborate its meaning. As for determining the polarity, AMR graphs are already equipped with this function through a polarity tag. Whether the polarity results in "not" being used depends on the length of the full constraint and the rule's position. For example, if only a single rule is used, the polarity tag indicates that a "not" should not be used due to the inverse nature of constraints. Once every rule has been handled, facts that indicate a relationship are created. An example of this would be a child's name and their gender. If these are specified within the structure of a fact, then they can be used to provide extra information within a constraint. These are handled in the same fashion as rules. This is also the point at which the format of the constraint is determined. A choice rule is used when a relational fact needs to be included and a variable remains unexplained. Doing so allows for a broader combination of possible results to be eliminated. There is a final check to ensure the polarities are as desired and to add any final formatting changes. Each constraint goes through this process until all are handled. Currently, there is nothing to handle multi-sentence constraints or ones that rely on information provided in different sections.

More complex problems, such as \textit{Einstein's Riddle}, as shown in Appendix Figure 2, require further consideration when formatting constraints. Keywords and tags in the AMR graph are used to determine when further formatting is required. For example, words such as "last" and "middle" indicate ordinality within a sequence. Combined with the fact that the order positions are contained, it is possible to clarify further what "middle" means within the context of the problem. If there are five houses, the middle node is replaced with "ordinal-entity 3". This allows the constraint to be generated as any other. Another type of positioning that requires altering the formatting occurs when the constraint refers to one object being "next to" or "to the right/left of" another. The resulting AMR contains the "location (next-to" or "relative-position" nodes, respectively. The individual objects are then defined, and the predetermined method of handling their relation is formatted. When one object is "to the right of" another, \texttt{O1 != O2 + 1} is used to modify the constraint appropriately. Further development of these knowledge modules or the production of another method is required as problems become more complex. Apart from these considerations, the process of solving this problem is the same as the \textit{Day-at-the-Zoo} example. As such, the constraints are simplified, and a subset is shown below: 

\begin{mdframed}\textit{\noindent
The Englishman's nationality is associated with the house color red.\\
The Spaniard's nationality is associated with the pet dog.\\
The beverage coffee is associated with the house color green.\\
The Ukrainian's nationality is associated with the beverage tea.
}
\end{mdframed}
%\newpage
The generated ASP from this subset is presented below:

\begin{mdframed}\textit{\noindent
:- not nationality\_of("red","englishman").\\ 
:- pet\_of(H,"dog"), not nationality\_of(H,"spaniard").\\
:- not beverage\_of("green","coffee").\\
:- beverage\_of(H,"tea"), not nationality\_of(H,"ukrainian").
}
\end{mdframed}
\newpage
\section{Results and Discussion}
Once fully assembled, the proposed system can convert unconstrained English to an equivalent ASP file. A section of the resulting constraints for the \textit{Day-at-the-Zoo} problem is shown in the following ASP code:

\begin{mdframed}\textit{\noindent
1\{order\_in\_line\_of(C,3):\\favorite\_animal\_of(C,"tigers"), gender\_of(C,"girl")\}1.\\
:- favorite\_animal\_of("Naomi","tigers").\\
1\{favorite\_animal\_of(C,"tigers"): \\not balloon\_design\_of(C,"hearts"),gender\_of(C,"girl")\}1.\\
1\{favorite\_animal\_of(C,"tigers"): \\not balloon\_design\_of(C,"stripes"),gender\_of(C,"girl")\}1.\\
:- order\_in\_line\_of("Johan",5).
}
\end{mdframed}

However, a correction had to be made to one of the lines. The constraint "Johan is not last in line" does not result in the correct AMR graph, as presented below:

\begin{mdframed}\textit{\noindent
\# ::snt Johan is not last in line.\\
(p / person\\
    \hspace*{3mm}:name (n / name\\
        \hspace*{6mm}:op1 "Johan")\\
    \hspace*{3mm}:ord (o / ordinal-entity\\
        \hspace*{6mm}:value -1)\\
    \hspace*{3mm}:ARG1-of (l / line-up-02))
}
\end{mdframed}

The performance of the AMR parser limits the efficacy of the proposed approach. If a mistake is made at this stage, no system is in place to correct it. In this case, the parser did not include a necessary polarity node. This results in the logical opposite of the desired constraint. While the AMR parser can produce incorrect results, these errors are relatively rare, which is shown by the fact that the AMR was only incorrect for one line in the entire problem description. This error has been corrected by hand so that the system's performance can be illustrated. Running the put-together \textit{Day-at-the-Zoo} ASP code results in the following output: 

\begin{mdframed}\textit{\noindent
solution("Kerry","tigers","swirls",3)\\ solution("Johan","chimpanzees","stripes",2)\\ solution("Mario","giraffes","hearts",1)\\ solution("Lani","lions","polka\_dots",5)\\ solution("Naomi","zebras","rainbow",4)
}
\end{mdframed}

The solution for \textit{Einstein's Riddle} results in the following output:

\begin{mdframed}\textit{\noindent
solution("englishman","red",3,"milk","old\_gold","snails") \\ solution("japanese","green",5,"coffee","parliaments","zebra")\\ solution("spaniard","ivory",4,"orange\_juice",\\"lucky\_strike","dog")\\ solution("norwegian","yellow",1,"water","kools","fox")\\ solution("ukrainian","blue",2,"tea","chesterfields","horse")
}
\end{mdframed}

When compared to the expected results, the outputs are correct. As previously noted, the example logic puzzles used are simpler ones. Further improvements to the current system are required to handle more complex problems. The possible inconsistencies within any stage utilizing an LLM are also worth noting. An LLM can provide different answers for each iteration given the same prompt. To help avoid this issue, smaller prompts with less complex demands are utilized. However, the possibility of an error is still present. These examples were able to produce the correct outcomes, but it is still worth noting.

\section{Conclusions and Future Work}
Answer Set Programming is a powerful declarative programming tool that allows for solving combinatorial problems. Generating ASP programs from unconstrained English would allow those with little to no knowledge to generate such programs. The system proposed in this paper takes a concrete step towards this goal by uniquely combining LLMs and AMRs for the first time in the context of solving logic puzzles. This approach allows the flexibility to handle unpredictable natural language inputs while maintaining structured semantic understanding, making errors more explainable. The former is what LLMs excel at handling and why they are used to gather important information from the problems. Moderating the use of LLMs and reducing the scope of each prompt helps avoid problems that arise from hallucinations and incorrect logic. AMRs allow for a point of predictability with their ability to consistently extract the meaning of sentences. The final result is a system capable of creating ASP files from unconstrained English. Future work on this system would involve expanding the capabilities of what can be generated and improving consistency. This would include, but is not limited to, implementing numerical comparisons, anonymous variables, and weighted rules. Doing so would allow for a broader scope of problems to be addressed. Certain logical concepts have yet to be addressed, such as handling more abstract positioning statements: "behind." Training an AMR that is more suited for the task at hand is also worth considering.

\bibliographystyle{IEEEtran}
\bibliography{references}
\newpage
\section*{Appendix}

\begin{mdframed}\textit{\noindent
Last Saturday, Kerry and four other children arrived early at the zoo so they could be the first ones in line (each child has a different favorite type of animal that he or she was anxious to see). As they waited in line, each child received a complimentary Mylar balloon with a different design on it. Two of the children - Johan and Mario - are boys, and three - Kerry, Lani, and Naomi - are girls. Can you determine the order (first, second, third, fourth, or fifth) each child stood in line, each child's favorite type of animal (one child likes the chimpanzees), and the design (one is a rainbow) on the balloon each received?
\begin{enumerate}
    \item The girl whose favorite animals are the tigers (who isn't Naomi) was third in line. Her balloon did not have hearts or stripes.
    \item Johan is not last in line and Mario is first in line. A boy is second in line. Lani's favorite animals are neither the tigers nor the zebras.
    \item Neither Lani's nor Naomi's balloon was either the one with hearts or the one with stripes. The balloon with stripes wasn't Mario's.
    \item The one whose favorite animals are the zebras, who isn't the one who received the balloon with swirls, was fourth in line. The one who received the balloon with polka dots likes the lions the most. Johan isn't the one whose favorite animals are the giraffes.
\end{enumerate}
}
\end{mdframed}
\centerline{\small{Appendix Figure 1: Day-at-the-Zoo Description}}
\vspace{1ex}

\begin{mdframed}\textit{\noindent
There are 5 houses in 5 different colors in a row, in a certain order. From left to right, the orders are 1, 2, 3, 4, and 5. In each house lives a person with a different nationality. The 5 owners drink a certain type of beverage, smoke a certain brand of cigar, and
keep a certain pet. No owners have the same pet, smoke the same brand of cigar, or drink the
same beverage. Now, who drinks water? Who owns the zebra?
\begin{enumerate}
    \item There are five houses.
    \item The Englishman lives in the red house.
    \item The Spaniard owns the dog.
    \item Coffee is drunk in the green house.
    \item The Ukrainian drinks tea.
    \item The green house is immediately to the right of the ivory house.
    \item The Old Gold smoker owns snails.
    \item Kools are smoked in the yellow house.
    \item Milk is drunk in the middle house.
    \item The Norwegian lives in the first house.
    \item The man who smokes Chesterfields lives in the house next to the man with the fox.
    \item Kools are smoked in the house next to the house where the horse is kept.
    \item The Lucky Strike smoker drinks orange juice.
    \item The Japanese smokes Parliaments.
    \item The Norwegian lives next to the blue house.
\end{enumerate}
}
\end{mdframed}
\centerline{\small{Appendix Figure 2: Einstein's Riddle Description}}

\end{document}